\documentclass[runningheads]{llncs}
\usepackage[T1]{fontenc}
\usepackage{graphicx}
\usepackage[hyphens]{url}
\usepackage[numbers,square,sort&compress]{natbib}
\usepackage{amsmath}
\usepackage{amssymb}
\usepackage{booktabs}
\usepackage{tabularx}
\usepackage{multirow}

\begin{document}
\title{CDBM: Complexity-Diversity Framework for Building Morphology Analysis}
\newcommand{\bench}{Shape-Bench}
\newcommand{\R}{\mathbb{R}}
\newcommand{\shapefeat}{\phi}

\newcommand{\numchar}{12}
\newcommand{\numtask}{33}
\newcommand{\numsingle}{12}
\newcommand{\numcase}{660}
\newcommand{\numrun}{3{,}300}
\newcommand{\numparcel}{227}
\newcommand{\numtool}{16}
\newcommand{\numloc}{990}
\newcommand{\corpusshort}{15.7}
\newcommand{\corpusfull}{15{,}730{,}508}
\newcommand{\trainrows}{12.2}

\author{Evgenii Maslov\inst{1} \and
Alexandra Vabnits\inst{1} \and
Denis Karpov\inst{1} \and
Alexey Morozov\inst{2} \and
Anastasia Antsiferova\inst{1} \and
Art Prosvetov\inst{1} \and
Valentin Khrulkov\inst{1}}
%
%
\institute{FusionBrain Lab, Artificial Intelligence Research Institute, 123112 Moscow, Russia \and
Independent Researcher}

\titlerunning{CDBM for Building Morphology Analysis}
\maketitle
\begin{abstract}
\emergencystretch=3em
We introduce Complexity–Diversity Building Metrics (CDBM), twelve interpretable descriptors for building massings represented as vertical extrusions. They measure how strongly geometric detail is expressed (complexity) and how unevenly it varies across one- and two-dimensional elements (diversity). Computed for 482,241,306 OpenStreetMap buildings and compared with 3D Building Metrics (3DBM) and the LION shape latent on task-specific samples, CDBM is tested on function, region, contextual relevance, real–generated discrimination, human preference and geometry-conditioned generation. With gradient boosting, CDBM best separates generated from mapped massings and best matches human preference; with logistic regression it matches 3DBM on realism. 3DBM leads on the other three tasks. As generation targets, the descriptors show that models struggle to produce strongly expressed and heterogeneous forms. CDBM complements existing indicators.

\end{abstract}
\section{Introduction}
\label{sec:intro}

Massing, the first volumetric statement of a building, fixes its footprint,
height and principal volumes on a site. Massings are increasingly generated
by rule-based procedures~\citep{wang2020massing} and by diffusion and
vision--language
models~\citep{rasoulzadeh2025archcomplete,jang2026footprint,maslov2026coma,sun2026knowledge}.
OpenStreetMap~\citep{haklay2008osm} and
GlobalBuildingAtlas~\citep{zhu2025globalbuildingatlas} databases map hundreds of
millions of buildings as footprints with heights, OpenStreetMap also with
building parts. Both can share one form, a set of vertical extrusions of
planar polygons: the LoD~1.3 (Level of Detail) block model~\citep{biljecki2016lod}.

Comparing massings requires metrics that describe their form. Urban
morphology uses building indicators to relate the built fabric to energy,
climate and other
phenomena~\citep{biljecki2022morphology,labetski2023metrics,fleischmann2019momepy}.
Generated massings are judged by humans, by task-specific metric
sets~\citep{alam2025sequential}, or by learned classifiers and distribution
distances~\citep{maslov2026coma,achlioptas2018learning,yang2019pointflow}
that return one score without saying which geometric property is missed. An
interpretable, cost-efficient descriptor defined on the generators' own
representation could analyse real buildings, diagnose generated ones and
specify geometric targets.

Existing representations meet these requirements only in part. The largest
inventories of building indicators are restricted to footprints, with height
as a single attribute~\citep{biljecki2022morphology}. True-3D
metrics~\citep{labetski2023metrics} compare the solid with a reference shape
and describe it as a whole: equal convexity may come from one deep recess or
from many small elements, spread over every part or concentrated in one wing.
General-purpose 3D descriptors and learned shape
latents~\citep{zeng2022lion,osada2002distributions,kazhdan2003spherical,novotni2004zernike}
accept arbitrary geometry, but operate on meshes, voxels or point samples
that no longer separate a building's contours from discretisation edges, and
their dimensions have no architectural meaning.

We introduce Complexity--Diversity Building Metrics (CDBM), twelve
interpretable descriptors defined directly on extruded-polygon massings. They
separate \emph{complexity}, how strongly a geometric property is expressed,
from \emph{diversity}, how unevenly it is distributed among the elements that
carry it, for three linear properties of the massing's skeleton (side length,
vertex angle and extrusion height) and three planar properties of its
footprints (circularity, convexity and elongation). Each descriptor is
dimensionless, bounded in $[0,1]$ and computed in closed form, so it scales to
hundreds of millions of buildings and can be requested from a generator as a
numerical target. CDBM complements existing indicators rather than replacing
them, allow interpretable comparison of existing generative methods and can help with improving massing generation.

Our contributions are:
\begin{itemize}
\item CDBM, twelve interpretable descriptors that separate complexity from
  diversity at the linear and planar levels (Section~\ref{sec:shape_descriptors}).
\item A comparison of CDBM with a domain-specific (3DBM~\citep{labetski2023metrics})
  and a learned (LION~\citep{zeng2022lion}) representation, computed on
  task-specific samples, on four classification tasks and on agreement with
  human preferences (Section~\ref{sec:experiments}).
\item An OpenStreetMap-derived dataset of 482 million LoD~1.3 buildings, all described by CDBM, and \corpusshort{} million sites with their massings (Section~\ref{sec:dataset}).
\item A benchmark of twelve single-descriptor generation tasks.
\end{itemize}
{\raggedright
We have released the code for the full experimental pipeline,
including the descriptors, at \url{https://huggingface.co/datasets/anonimous-submission/CDBM}, and the dataset at
\url{https://huggingface.co/datasets/anonimous-submission/OSM}.
\par}

\section{Related Work}
\label{sec:related}

\subsection{3D Representation Learning}

Learned 3D representations range from voxel-based energy models
and continuous implicit fields~\citep{xie2018descriptor,
mescheder2019occupancy} to hierarchical and spatially structured
latents for diffusion-based generation~\citep{zeng2022lion,
xiang2024trellis}. Architectural applications use variational
autoencoders for building-type interpolation and sketch-based
reconstruction~\citep{demiguel2020interpolations,tono2024vitruvio},
while sequential and hierarchical models support massing
preference evaluation, autocompletion and architectural
generation~\citep{alam2025sequential,rasoulzadeh2025archcomplete}.
Although these methods encode building geometry, their learned
features generally lack explicit geometric meaning, motivating
dedicated indicators for interpretable
morphological analysis.

\subsection{Building morphology indicators}

General 3D shape descriptors use sampled geometric
distributions~\citep{osada2002distributions}, spherical
harmonics~\citep{kazhdan2003spherical} and Zernike
moments~\citep{novotni2004zernike} to encode overall
shape for comparison and retrieval.
Urban morphology indicators characterize spatial
organization, street networks, land use and
building--plot relationships~\citep{zhang2023spatial}.
Building-scale indicators quantify footprint compactness,
rectangularity and vertex counts, supplemented by
contextual measures and height-derived
properties~\citep{biljecki2022morphology}.
Three-dimensional indicators capture urban height
and canopy structure alongside building volume,
surface area, convexity, cuboidness and
roughness~\citep{li2024parameters,labetski2023metrics}.
These approaches primarily describe planar geometry
or aggregate spatial and volumetric properties,
leaving the detail and heterogeneity of individual
building parts less explicitly characterized.

\subsection{Building Layout Generation}

Massing generation methods include procedural, optimization-based,
learning-based and multimodal approaches.
Procedural systems use rules and additive--subtractive
operations~\citep{wang2020massing}, while optimization-based
methods combine generation and simulation to satisfy
environmental objectives~\citep{huang2025performance,
bouillon2025multilod,wang2024evomass,feng2024morphology}.
Learning-based methods employ graph-conditioned
GANs~\citep{chang2021building}, graph and vector models
for building layouts~\citep{xu2021blockplanner,he2023globalmapper},
and variational autoencoders and diffusion-based models
for architectural geometry~\citep{tono2024vitruvio,
rasoulzadeh2025archcomplete,jang2026footprint}.
Multimodal approaches combine text, site geometry and
contextual imagery through knowledge-augmented
workflows~\citep{sun2026knowledge}, vision--language
models~\citep{maslov2026coma} and context-informed
diffusion~\citep{zhou2026controlcity}.

\subsection{3D Urban Datasets}

Architectural datasets include Manhattan1K, with 1{,}000
building models and synthetic sketches for
reconstruction~\citep{tono2024vitruvio}, and the 1{,}500
voxelized house models used by
ArchComplete~\citep{rasoulzadeh2025archcomplete}.
BuildingWorld provides approximately five million LoD2
models and LiDAR point clouds for reconstruction,
detection and segmentation~\citep{huang2026buildingworld}.
Structured LoD~1.3 representations preserve building
parts with differing heights~\citep{maslov2026coma,
biljecki2016lod}.
At global scale, GlobalBuildingAtlas provides footprints,
heights and LoD1 models for 2.68 billion
buildings~\citep{zhu2025globalbuildingatlas}, but its
single-extrusion representation omits variation among
volumetric parts, limiting detailed analysis of
within-building morphology.

\section{CDBM Framework}
\label{sec:shape_descriptors}

We adopt the hierarchical massing representation introduced
in CoMa~\citep{maslov2026coma}, corresponding to the LoD~1.3
abstraction of vertically differentiated building
parts~\citep{biljecki2016lod}.
A massing describes the overall volumetric form of one
or more buildings as collections of vertical polygonal
extrusions:
\begin{equation}
\label{eq:massing}
m=(b_1,\dots,b_K),\quad
b_k=(e_{k,1},\dots,e_{k,L_k}),\quad
e_{k,l}=
\bigl(z^\downarrow_{k,l},z^\uparrow_{k,l},
(P_{k,l,1},\dots)\bigr),
\end{equation}
where $K$ is the number of buildings, $L_k$ the number
of extrusions in building $b_k$, and each extrusion
$e_{k,l}$ is defined by its bottom and top elevations
and footprint polygons.
Each polygon is an ordered sequence of planar vertices:
\begin{equation}
\label{eq:polygon}
P_{k,l,r}=(p_{k,l,r,1},\dots,p_{k,l,r,N_{k,l,r}}),
\qquad
p_{k,l,r,i}=(x_{k,l,r,i},y_{k,l,r,i})\in\mathbb{R}^2,
\end{equation}
where $r$ indexes the polygons of an extrusion and
$N_{k,l,r}$ denotes the number of vertices.
We serialize this hierarchy using the same JSON schema
as CoMa.

\subsection{Two axes of geometric complexity}
\label{sec:descriptor-axes}

We describe building geometry through twelve complementary
morphological indicators, organized along two axes:
complexity versus diversity, and linear versus planar
properties.

The first axis is \emph{complexity versus diversity}, and it is our account of geometric detailing. Complexity measures how strongly a property is expressed: short edges, skewed corners, intermediate slabs, non-circular, concave or elongated footprints. Diversity measures how unevenly that property is distributed among the elements that carry it. High diversity means a mix of fine and coarse detail---long walls with short jogs, square corners with irregular ones, full-height volumes beside shallow setbacks. Complexity therefore captures the fineness of detailing; diversity captures its heterogeneity.

The second axis is \emph{linear versus planar}. We ask the same two questions at successive geometric levels, starting from the one-dimensional skeleton of the massing and continuing to its two-dimensional footprints. Linear scores act on edges, vertices and extrusion heights. Planar scores act on the shape of each footprint. A dedicated three-dimensional level is omitted: under LoD~1.3, volumetric features such as setbacks, courtyards and stepped silhouettes are already the result of stacking polygonal footprints at different heights. A 3D convexity or circularity of the solid would largely recombine the same linear and planar signals rather than add an independent view of complexity.

\begin{figure}[h]
\includegraphics[width=\textwidth]{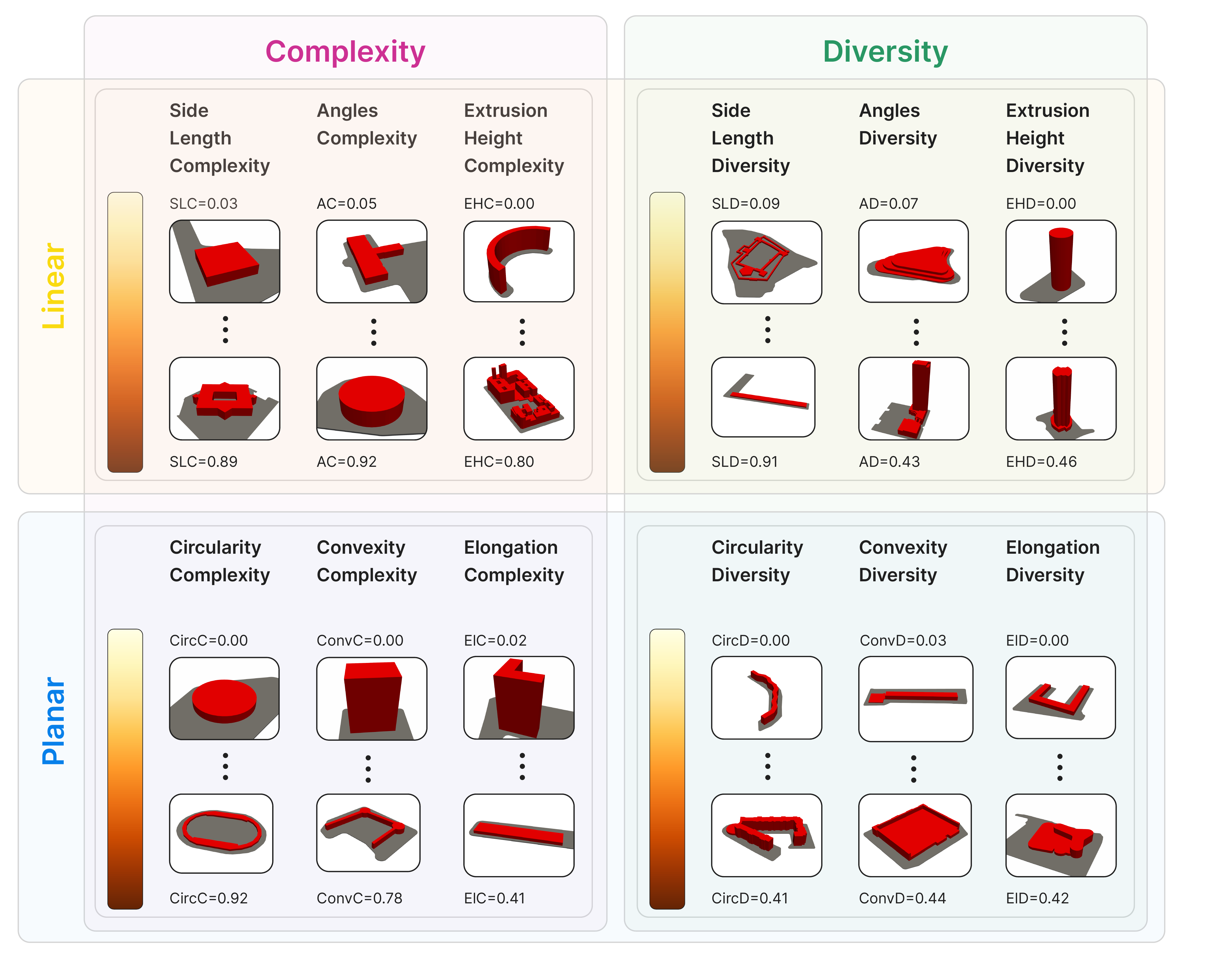}
\caption{Geometric interpretation of the twelve CDBM descriptors, organized by complexity versus diversity (columns) and linear versus planar properties (rows). Each descriptor is illustrated by massings with lower and higher values, arranged from top to bottom; the corresponding scores are shown.}
\label{fig:shape_descriptors}
\end{figure}

Crossing these axes yields four groups: linear complexity,
linear diversity, planar complexity, and planar diversity.
Each group contains three distinct metrics, giving twelve
descriptors normalized to $[0,1]$. Together, they characterize
the expression and heterogeneity of geometric properties
at different levels of building form.
Section~\ref{sec:descriptor-defs} defines the descriptors,
and Figure~\ref{fig:shape_descriptors} illustrates their
geometric meaning.

\subsection{Metric Definitions}
\label{sec:descriptor-defs}

Each metric first assigns scores to geometric elements
of a building and then aggregates them using the mean
$\mu$ for complexity or the population standard deviation
$\sigma$ for diversity. For an indexed multiset $R$,
\begin{equation}
\label{eq:mean-std}
\mu(R)=\frac{1}{|R|}\sum_{r\in R}r,
\qquad
\sigma(R)=
\sqrt{\frac{1}{|R|}\sum_{r\in R}
\bigl(r-\mu(R)\bigr)^2}.
\end{equation}
For massings containing several buildings, complexity
pools the corresponding scores across buildings,
while diversity averages their within-building
standard deviations.

For building $b_k$, edge length $\ell_{k,l,r,i}$
and vertex angle $\theta_{k,l,r,i}$ refer to polygon
$P_{k,l,r}$, with cyclic vertex indexing.
Angles are the smaller angles between adjacent edges.
Lengths are rounded to two decimal places and angles
to the nearest degree before scoring.
$\Lambda_k$ denotes the longer side of the minimum-area
bounding rectangle of the union of the building's
extrusion footprints.
Extrusion height is
$h_{k,l}=z^\uparrow_{k,l}-z^\downarrow_{k,l}$,
and full building height is
$H_k=\max_l z^\uparrow_{k,l}-\min_l z^\downarrow_{k,l}$.
For each extrusion footprint, $A_{k,l}$ is its area,
$\Pi_{k,l}$ its perimeter,
$A^{\mathrm{conv}}_{k,l}$ its convex-hull area,
and $a_{k,l}\le b_{k,l}$ the side lengths of its
minimum-area bounding rectangle.
The following indexed multisets retain one score
per contour edge, contour vertex, or extrusion;
repeated values are retained.
$\mathbf{1}$ denotes the indicator function.

\begin{equation}
\label{eq:sidelen}
R_{\mathrm{slc}}
=
\left\{
\max\!\left(
1-\frac{\ell_{k,l,r,i}}{\Lambda_k},\,0
\right)
\right\}_{l,r,i},
\qquad
\text{side length}.
\end{equation}

\begin{equation}
\label{eq:angles}
R_{\mathrm{ac}}
=
\left\{
\frac{|\theta_{k,l,r,i}-90^\circ|}{90^\circ}
\mathbf{1}_{\theta_{k,l,r,i}\ne180^\circ}
\right\}_{l,r,i},
\qquad
\text{angle}.
\end{equation}

\begin{equation}
\label{eq:height}
R_{\mathrm{ehc}}
=
\left\{
\max\!\left(1-\frac{h_{k,l}}{H_k},\,0\right)
\right\}_{l=1}^{L_k},
\qquad
\text{extrusion height}.
\end{equation}

\begin{equation}
\label{eq:circularity}
R_{\mathrm{circ}}
=
\left\{
1-\frac{4\pi A_{k,l}}{\Pi_{k,l}^{2}}
\right\}_{l=1}^{L_k},
\qquad
\text{circularity}.
\end{equation}

\begin{equation}
\label{eq:convexity}
R_{\mathrm{conv}}
=
\left\{
1-\frac{A_{k,l}}{A^{\mathrm{conv}}_{k,l}}
\right\}_{l=1}^{L_k},
\qquad
\text{convexity}.
\end{equation}

\begin{equation}
\label{eq:elongation}
R_{\mathrm{el}}
=
\left\{
1-\frac{a_{k,l}}{b_{k,l}}
\right\}_{l=1}^{L_k},
\qquad
\text{elongation}.
\end{equation}

Table~\ref{tab:descriptor_definitions} defines the twelve
building-level metrics by applying $\mu$ and $\sigma$
to these multisets and summarizes their geometric meaning.

\begin{table}[t]
\centering
\small
\setlength{\tabcolsep}{4pt}
\renewcommand{\arraystretch}{1.1}
\caption{Building-level descriptor definitions.
Interpretations describe higher metric values.}
\label{tab:descriptor_definitions}
\begin{tabularx}{\linewidth}{
@{}p{0.25\linewidth} l l X@{}
}
\toprule
Metric & Abbr. & Formula & Interpretation \\
\midrule

Side length complexity
& SLC
& $\mu(R_{\mathrm{slc}})$
& Shorter contour edges relative to building extent. \\

Side length diversity
& SLD
& $\sigma(R_{\mathrm{slc}})$
& Greater variation in relative contour edge lengths. \\

Angle complexity
& AC
& $\mu(R_{\mathrm{ac}})$
& Stronger departure from orthogonal geometry. \\

Angle diversity
& AD
& $\sigma(R_{\mathrm{ac}})$
& Greater variation in angular departure from
orthogonal geometry. \\

Extrusion height complexity
& EHC
& $\mu(R_{\mathrm{ehc}})$
& Shorter extrusion heights relative to full
building height. \\

Extrusion height diversity
& EHD
& $\sigma(R_{\mathrm{ehc}})$
& Greater variation in relative extrusion heights. \\

Circularity complexity
& CirC
& $\mu(R_{\mathrm{circ}})$
& Stronger departure from circular compactness. \\

Circularity diversity
& CirD
& $\sigma(R_{\mathrm{circ}})$
& Greater variation in circularity across
extrusion footprints. \\

Convexity complexity
& ConC
& $\mu(R_{\mathrm{conv}})$
& More pronounced footprint concavity. \\

Convexity diversity
& ConD
& $\sigma(R_{\mathrm{conv}})$
& Greater variation in convexity across
extrusion footprints. \\

Elongation complexity
& ElC
& $\mu(R_{\mathrm{el}})$
& More elongated extrusion footprints. \\

Elongation diversity
& ElD
& $\sigma(R_{\mathrm{el}})$
& Greater variation in elongation across
extrusion footprints. \\

\bottomrule
\end{tabularx}
\end{table}

\section{Experiments}
\label{sec:experiments}

\subsection{Data Sources}
\label{sec:dataset}

We construct a building dataset from 990 regional
OpenStreetMap (OSM) extracts downloaded on
1 June 2026~\citep{osmextracts}, comprising
482{,}241{,}306 buildings across various regions
and functional categories.
A custom pipeline reconstructs buildings and their parts
in the representation defined in
Section~\ref{sec:shape_descriptors}, retaining contour
detail and vertical differentiation while resolving
overlapping volumes and checking geometric validity.
The algorithm is described in the released code at \url{https://huggingface.co/datasets/anonimous-submission/CDBM}.
We additionally assemble 15{,}730{,}508 sites with their
associated building massings by polygonizing ground-level
road networks and mapped area boundaries, refining the
resulting polygons, and spatially matching them to buildings.

\begin{figure}[h]
\includegraphics[width=\textwidth]{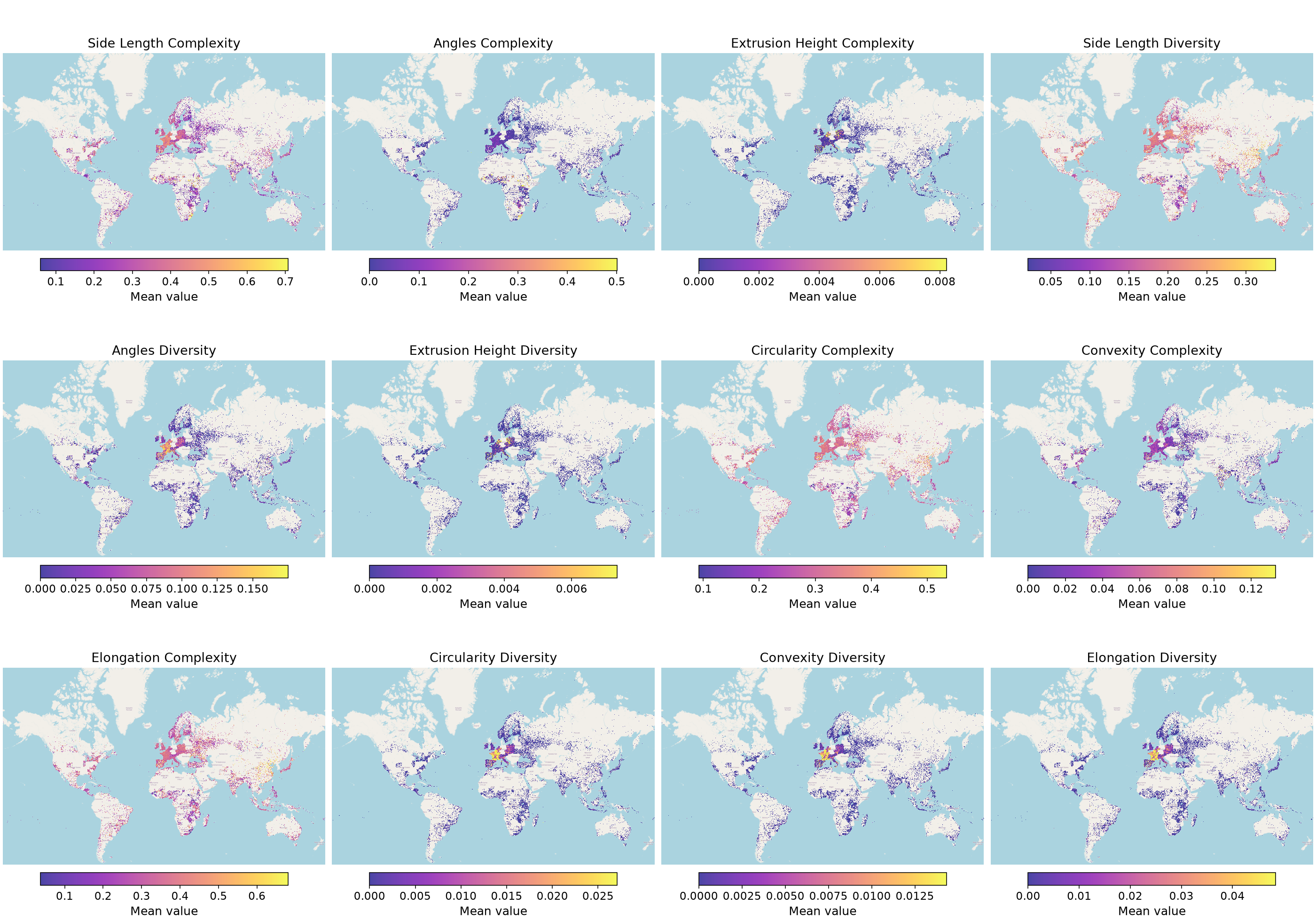}
\caption{Geographic distribution of the twelve CDBM descriptors in the OSM-derived building dataset. Each occupied grid cell shows the mean descriptor value of its buildings. Colour scales are specific to each descriptor.}
\label{fig:descriptor_maps}
\end{figure}

The proposed site--massing dataset supplies the training and
validation data for the classification and generative
experiments described below.
These experiments use only single-building sites,
although the dataset also contains multi-building sites
supporting future generation tasks.
The dataset constitutes an additional contribution
of this work.
Figure~\ref{fig:descriptor_maps} shows its geographic
coverage and descriptor distributions, with maps cropped
to the extent of the observed buildings.
Each occupied cell of a fine spatial grid displays
the mean descriptor value of its buildings.
Coverage gaps reflect the use of selected regional
extracts containing only part of the global OSM data.

\subsection{Baselines}
\label{sec:baselines}

We compare our descriptors with two external representations of 
building massing: a domain-specific set of 
morphological metrics, 3D Building Metrics (3DBM) 
~\citep{labetski2023metrics}, and a learned, general-purpose 
shape representation, the latent space of the Latent Point 
Diffusion Model (LION)~\citep{zeng2022lion}. We use both as 
released, without re-implementing 3DBM or fine-tuning LION. 

\subsubsection{3D Building Metrics}

3DBM computes 75 values per building, notably size-independent
2D and 3D shape indices that compare the footprint or the solid
with a reference shape of equal area or volume
~\citep{basaraner2017indices, fleischmann2019momepy}. Of the 75 output values, we use the 
42 in Table~\ref{tab:3dbm}. We drop the other 33: six bookkeeping 
fields; 10 vertex and face counts, which describe the 
triangulation rather than the shape; 9 redundant or unstable 
values, such as the convex-hull volume, already contained in 
convexity; four values that depend on the site's position and 
orientation in its coordinate system; two neighbourhood metrics, 
which are undefined for an isolated site; and the exchange index, 
which is absent from the results of~\citep{labetski2023metrics}.

\begin{table}[h]
\caption{3DBM values used as features. $^\ast$Recomputed 
by us, see deviations (i)--(ii).}
\label{tab:3dbm}
\centering\scriptsize\setlength{\tabcolsep}{4pt}
\begin{tabular}{@{}lrp{0.85\textwidth}@{}}
\toprule
Group & \# & Values \\
\midrule
3D    & 18 & hemisphericality, convexity, fractality, cuboidness, 
             cubeness, proximity, spin, circumference, depth, 
             girth, dispersion, range, equivalent cuboid, roughness; 
             horizontal elongation; min/max vertical elongation$^\ast$; 
             form factor$^\ast$ \\
2D    & 14 & the 2D analogues of these 14 indices, computed on the 
             footprint \\
size  & 10 & volume, surface area, footprint area and perimeter, 
             bounding-box length and width; roof height: min, max, 
             mean, std \\
\bottomrule
\end{tabular}
\end{table}

\subsubsection{LION}
LION~\citep{zeng2022lion} is a generative model of 3D shapes 
represented as point clouds. The encoder maps a point 
cloud of 2{,}048 points to a 128-dimensional global latent $z_0$ 
and, given it, to $2048 \times 4$ point latents $h_0$. We use the 
encoder trained on 55 ShapeNet categories~\citep{chang2015shapenet,yang2019pointflow}, 
frozen\footnote{\url{https://github.com/nv-tlabs/LION}}: 
buildings are rare in its data, but fine-tuning on our 
labels would tie it to the evaluated task.

\subsection{Real-Generated Classification and Human Preference Alignment}
\label{sec:fake_buildings_detection}

We test whether the proposed descriptors distinguish
OSM-derived from AI-generated massings and whether
classifier scores align with human visual preferences,
comparing our descriptors with baseline features.

\paragraph{Data and generators.}
We sample 811{,}023 single-building training sites from our dataset
(Section~\ref{sec:dataset}), balancing the marginal distributions of
descriptors, building area, floor count, site geometry, function, and region, and hold out another 1{,}100 sites for generation and evaluation.
We fine-tune Qwen3-VL-8B-Instruct with
LoRA~\citep{bai2025qwen3vl,hu2022lora} and train a
TRELLIS-based generator~\citep{xiang2024trellis}
on this split, subject to model-specific preprocessing.
We also evaluate Qwen3.5-35B-A3B and Qwen3.6-35B-A3B
zero-shot. Each method generates one massing per held-out
site, conditioned on its boundary, building function,
area, and floor count.
Outputs use the polygon-extrusion JSON representation.

\paragraph{Real--generated classification.}
We use 1{,}000 held-out sites to assemble 1{,}000 reference
and 1{,}000 generated massings, with one generation
per site and 250 per method.
For each feature representation, CatBoost~\citep{prokhorenkova2018catboost} and logistic
regression~\citep{cox1958regression} are evaluated using identical grouped
$k$-fold splits, keeping reference and generated massings
from the same site together.
We report mean performance across folds.

\paragraph{Alignment with human preferences.}
The remaining 100 sites contribute 25 generated massings
per method. Four annotators evaluate 2{,}500 pairs side by side,
matched by area and floor count, with coverage of all generators,
selecting the better massing or indicating that the two are equal.
A Bradley--Terry model converts pairwise preferences
into individual scores.
Each classifier is then retrained on all 2{,}000
classification examples and applied to these 100 massings.
We use the predicted probability of the OSM-derived
class as a realism score and report its Spearman and
Pearson correlations with human preference scores.
The preference set is excluded from classifier training
and model selection.

\subsection{Contextual Relevance}
\label{sec:context_setup}

A massing should also fit its surroundings. CoMa~\citep{maslov2026coma} scores
this \emph{contextual relevance} with a classifier on 17 hand-crafted features
that compare the massing with its neighbours. We ask whether per-building shape
descriptors also carry contextual information and add to these features.

\begin{figure}[tb]
\centering
\includegraphics[width=0.7\textwidth]{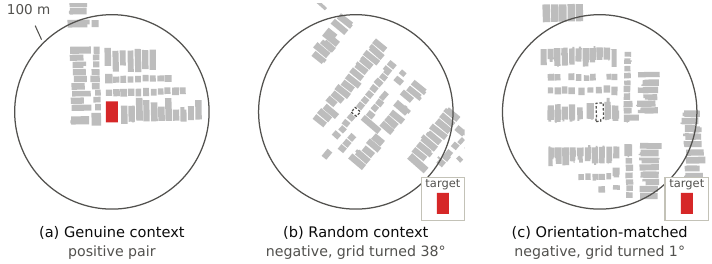}
\caption{A target (red) in Chicago with three contexts (grey, buildings
within 100\,m): (a)~the true context; (b)~a random false context, given away
by a street grid turned by $38^{\circ}$; (c)~an orientation-matched false
context. Insets show the target at the same scale; dashed outlines mark the
foreign site's own building.}
\label{fig:context_task}
\end{figure}

\paragraph{Task.}
To test this, we train a classifier to tell a massing's true context from a
false one (Fig.~\ref{fig:context_task}). In continuous spatial data, a pair of
a target massing and a context (a set of surrounding buildings) is
\emph{true} (positive) if the context surrounds the target and \emph{false}
(negative) if it comes from another site; the predicted probability of a true
pair is the contextual relevance score. Since every target appears in both
kinds of pairs, its own descriptors score 0.5 ROC-AUC: the signal lies in how
it compares with its context. Following CoMa, we compute each descriptor $d$
on the target $m$ and the context buildings $c_1,\dots,c_n$ (mean $\mu$,
standard deviation $\sigma$) and aggregate it in two ways:
\begin{equation}
\label{eq:context_kernel}
f_{\mathrm{mean}}=\exp\!\left(-\frac{|\mu-d(m)|}{\sigma}\right),
\qquad
f_{\mathrm{near}}=\exp\!\left(-\frac{\min_{j}|d(c_j)-d(m)|}{\sigma}\right),
\end{equation}
high when the target is typical of its context ($f_{\mathrm{mean}}$) or when
the context contains a similar building ($f_{\mathrm{near}}$). Each descriptor
thus yields two features, and all descriptor sets are compared on this doubled
representation.

\paragraph{Descriptor sets.}
Besides the baselines of Sec.~\ref{sec:baselines}, 3DBM and LION, we compare with
CoMa's 17 features (CoMa-17), which already compare target and context and are
used as is. Here LION is made orientation-invariant: we align each footprint's
long axis with the $x$-axis and average the latents over its symmetric turns,
since its raw latents also encode the building's orientation in the world.
To see which properties drive contextual fit, we split CoMa-17 and
3DBM into semantic families, giving eight families in four groups
(Table~\ref{tab:context_families}).

\begin{table}[tb]
\caption{Groups and families compared on contextual relevance. \#: features;
per-building descriptors count twice (Eq.~\ref{eq:context_kernel}).}
\label{tab:context_families}
\centering\scriptsize\setlength{\tabcolsep}{4pt}
\begin{tabular}{@{}llrp{0.62\textwidth}@{}}
\toprule
Group & Family & \# & Features \\
\midrule
LION    & -- & 288 & global latent (128) and pooled point latents (16) \\
\midrule
3DBM    & 3D shape & 36 & the 18 3D indices of Table~\ref{tab:3dbm} \\
        & 2D shape & 28 & the 14 2D indices \\
        & size     & 20 & the 10 size values \\
\midrule
CoMa & orientation & 5 & orientation difference to the context mean and the nearest
  building; orientation density; two distribution distances (orientation;
  orientation, size and shape) \\
        & shape & 7 & circularity, elongation, courtyard and stepback ratios; diversity
  of angles, extrusion heights and elevations \\
        & site and size & 5 & floor-area ratio, coverage, setback; height (mean, nearest) \\
\midrule
CDBM (ours) & -- & 24 & the twelve complexity and diversity descriptors \\
\bottomrule
\end{tabular}
\end{table}

\paragraph{Benchmark.}
To go beyond CoMa's single city and random negatives, we build a benchmark from our dataset (Section~\ref{sec:dataset}), comprising 15{,}987 targets from the densest 4\,km$^2$ of 22 cities (19 countries, six continents), approximating each site by a morphological tessellation
cell~\citep{fleischmann2020tessellation}. Since the score depends
on what counts as context and how false contexts are drawn, we vary both, and
the test split:
\begin{itemize}
\item \emph{Context}: all buildings within 100\,m (as in CoMa) or the 50
  nearest within 300\,m.
\item \emph{False context}: from a random same-city site (as in CoMa); from
  the most similarly oriented of 64 such sites (\emph{orientation-matched}),
  removing the easiest cue, a turned street grid
  (Fig.~\ref{fig:context_task}b); or from another city (\emph{cross-city}).
  The foreign site's own building is excluded.
\item \emph{Test set}: held-out sites or seven held-out cities.
\end{itemize}
This gives 12 classification tasks (\emph{settings}).

\paragraph{Evaluation protocol.}
In each of the 12 settings (three kinds of negatives, two contexts, two test
sets), we train a CatBoost classifier on
each of the 255 non-empty combinations of the eight families and score it by
ROC-AUC on the test set. The eight settings with random or
orientation-matched negatives are the \emph{main settings}, over which we
average; cross-city negatives are nearly trivial (0.98 for CoMa-17) and
excluded from the averages.
Since many differences are a few thousandths, two sets are compared with
paired 95\% bootstrap intervals over the test targets. To check that our
conclusions about which families matter do not rest on the test data, we also
run a forward selection over families on the training data alone.

\subsection{Geographic Origin and Building Function}
\label{sec:classification_methods}

We investigate whether CDBM captures morphological patterns
associated with building function and geographic origin.
We construct eleven classification tasks from the
single-building sites in the dataset described in
Section~\ref{sec:dataset}, using its regional organization
to select data, apply sampling rules, and form training
and test subsets.

\noindent\textit{Geographic classification.}
The dataset preserves the regional hierarchy of the
source OSM extracts~\citep{osmextracts}.
We define three groups of geographic classes at different
scales: Global8 groups sources into eight world regions,
Urban8 covers eight major city agglomerations across different
continents, and Europe15 covers fifteen European countries.
Their constituent regions and sampling configurations
are listed in Table~\ref{tab:classification_geographic_tasks}.
Urban8 includes the selected cities and neighboring
territories within 100~km of their centers.

Each geographic class contains source subregions,
defined here as leaf-level locations in the extract
hierarchy. Because the sampled subregions can affect
the observed morphology, we compare three sampling
schemes. \emph{Random sampling} selects sites within
each class without subregional quotas, approximately
preserving the relative contributions of its subregions.
\emph{Location-capped sampling} limits the candidate
pool to 1{,}000 sites per subregion, reducing domination
by the largest sources.
\emph{Territory-balanced sampling} allocates candidate
quotas equally across subregions and reallocates
unfilled quotas to those with remaining eligible sites;
the achievable balance is constrained by the number
of subregions and their available data.

\begin{table}[t]
\caption{Geographic classification tasks.
Counts refer to complete datasets before the
training/test split. Subregions are leaf-level
locations in the OSM extract hierarchy.}
\label{tab:classification_geographic_tasks}
\centering
\scriptsize
\setlength{\tabcolsep}{4pt}
\renewcommand{\arraystretch}{1.25}
\begin{tabularx}{\linewidth}{
@{}p{0.11\textwidth}p{0.48\textwidth}Xr@{}
}
\hline
Group & Regions / classes & Sampling & Total sites \\
\hline

\multirow{2}{=}{Global8}
& \multirow{2}{=}{\raggedright
Africa, Asia, Central America, Europe,
North America, Oceania, Russia, South America}
& Location-capped\newline
(8{,}000 per region)
& 64{,}000 \\
& & Random\newline
(8{,}000 per region)
& 64{,}000 \\

\hline
Urban8
& \raggedright Moscow, London, Paris, New York,
Mexico City, Beijing, Shanghai, Delhi
& Random\newline
(5{,}000 per urban area)
& 40{,}000 \\

\hline
\multirow{2}{=}{Europe15}
& \multirow{2}{=}{\raggedright
France, Spain, Ukraine, Switzerland, Sweden,
Italy, Portugal, Finland, Norway, Poland,
the Netherlands, Czech Republic,
the United Kingdom, Austria, Belgium}
& Territory-balanced\newline
(8{,}000 per country)
& 120{,}000 \\
& & Random\newline
(8{,}000 per country)
& 120{,}000 \\

\hline
\end{tabularx}
\end{table}

\noindent\textit{Building-function classification.}
We construct six tasks using Moscow, the pooled Urban8
areas, Calgary, and Shanghai.
Each task predicts one of seven functions:
apartment buildings, individual houses, retail,
education, industrial, office, and healthcare.

Function-class availability varies across sources,
so we compare three sampling schemes.
\emph{All eligible} uses all eligible candidates,
retaining up to 8{,}000 sites per function in Moscow
or 1{,}000 per urban-area--function combination
in Urban8; smaller groups are retained in full,
leaving the function distribution imbalanced.
\emph{Class-balanced sampling} selects 700 sites
per function from Urban8 without balancing urban-area
contributions within functions.
\emph{Random sampling} selects 1{,}800 sites per city
without function-class quotas.
Table~\ref{tab:classification_function_tasks}
lists the resulting source--sampling combinations.

\begin{table}[t]
\caption{Building-function classification tasks.
Counts refer to complete datasets before the
training/test split.
The Moscow ``all eligible'' sample contains
8{,}000 apartment buildings and 2{,}567 buildings
from the other six classes.}
\label{tab:classification_function_tasks}
\centering
\scriptsize
\setlength{\tabcolsep}{4pt}
\begin{tabular*}{\linewidth}{
@{\extracolsep{\fill}}llr@{}
}
\toprule
Source & Sampling & Total sites \\
\midrule
\multirow{2}{*}{Moscow}
& All eligible & 10{,}567 \\
& Random & 1{,}800 \\
\midrule
\multirow{2}{*}{Urban8}
& All eligible & 23{,}322 \\
& Class-balanced & 4{,}900 \\
\midrule
Calgary & Random & 1{,}800 \\
\midrule
Shanghai & Random & 1{,}800 \\
\bottomrule
\end{tabular*}
\end{table}

Across the eleven tasks, there are 452{,}189 observations
and 397{,}813 unique sites, reflecting overlap between
task datasets.

\noindent\textit{Features and comparison protocol.}
We compare CDBM (12 features) with LION (144 features)
and the selected 3DBM representation (42 features),
described in Section~\ref{sec:baselines}.
To assess information beyond building scale,
we evaluate each representation with and without
\emph{Base}, consisting of gross floor area and
floor count.
For geographic classification, the reference set is
\emph{Base+Function}, which additionally includes
building function as a categorical feature.
In both Urban8 building-function tasks, urban area
is supplied to every feature configuration.

For each task, we evaluate seven input sets:
the reference set $B$, the three representations
$X\in\{\mathrm{CDBM},\mathrm{LION},\mathrm{3DBM}\}$,
and their combinations $B\cup X$.
Here, $B$ is Base for function classification
and Base+Function for geographic classification.
Each input set is evaluated with CatBoost and
logistic regression, yielding 154 task--model--feature
combinations.

The primary metric is test macro-averaged one-vs-rest
ROC--AUC, reported on the $[0,1]$ scale.
The improvement over the reference set is
\begin{equation}
\Delta_X
=
\operatorname{AUC}(B\cup X)
-
\operatorname{AUC}(B).
\end{equation}

\subsection{Geometry-Conditioned Generation Benchmarking}
\label{sec:shape_bench}

We propose a descriptor-based evaluation protocol for assessing
how well generative models follow different geometric concepts
and produce massings with varying levels of geometric complexity
and diversity. Each proposed descriptor defines an explicit
generation target, allowing us to evaluate geometric control
through agreement between the requested value and the value
measured on the generated massing.

\paragraph{Generative methods.}
We evaluate the same zero-shot models as in
Section~\ref{sec:fake_buildings_detection}, together with
descriptor-conditioned versions of the two trained generators.
These versions use the same architectures, training data,
and optimization procedures as in the preceding experiment.
The only change is the addition of shape descriptor targets
to the conditioning inputs: during training, each example
includes a randomly selected subset of descriptor values
computed from its reference massing, alongside the building
function, area, floor count, and site boundary.
The number of supplied descriptors varies between examples,
exposing the models to geometric requirements with different
degrees of specification.

\paragraph{Benchmark construction.}
The benchmark uses the same 1{,}000 held-out sites as the
real--generated classification experiment
(Section~\ref{sec:fake_buildings_detection}).
For each site, we compute all twelve descriptors of the
reference massing and construct twelve generation instances.
Each instance retains the site and building requirements
and adds the target value of exactly one descriptor.
This yields twelve tasks with 1{,}000 instances each,
covering the complexity and diversity variants of the six
geometric properties defined in
Section~\ref{sec:shape_descriptors}.
The resulting 12{,}000 instances evaluate each descriptor
in isolation, with target values drawn from observed
building geometries.

\paragraph{Evaluation protocol.}
Each model is run five times on every benchmark instance,
giving 60{,}000 generation runs per method.
For a target descriptor value $d \in [0,1]$ and its measured
value $\hat{d} \in [0,1]$ in the generated massing, we define
the target-match score as
\begin{equation}
\label{eq:descriptor_target_match}
s(d,\hat{d}) = 1 - |d-\hat{d}|.
\end{equation}
The score ranges from zero to one, with one indicating
an exact match. For each instance $i$ and descriptor $j$,
we retain the highest score across the five runs,
\begin{equation}
\label{eq:descriptor_best_of_five}
s^{*}_{i,j}
=
\max_{r \in \{1,\ldots,5\}}
\left(1-\left|d_{i,j}-\hat{d}_{i,j}^{(r)}\right|\right).
\end{equation}
For each descriptor, we report the mean best-of-five score
over the 1{,}000 evaluation instances and examine performance
across target values. This measures the ability to satisfy
each geometric requirement within a fixed budget of five
generation attempts.

\section{Results}
\label{sec:results}

\subsection{Real--Generated Classification and Human Preference Alignment}
\label{sec:preference_results}

\begin{table}[t]
\centering
\caption{Real--generated classification performance
(ROC--AUC) for each feature representation.
Best results for each classifier are shown in bold.}
\label{tab:classifiers_results}
\begin{tabular*}{\linewidth}{
@{\extracolsep{\fill}}lcc@{}
}
\toprule
Features & Logistic Regression & CatBoost \\
\midrule
LION & 0.810 & 0.837 \\
3DBM & 0.861 & 0.931 \\
CDBM (ours) & \textbf{0.862} & \textbf{0.972} \\
\bottomrule
\end{tabular*}
\end{table}

\begin{figure}[t]
\centering
\includegraphics[width=\linewidth]{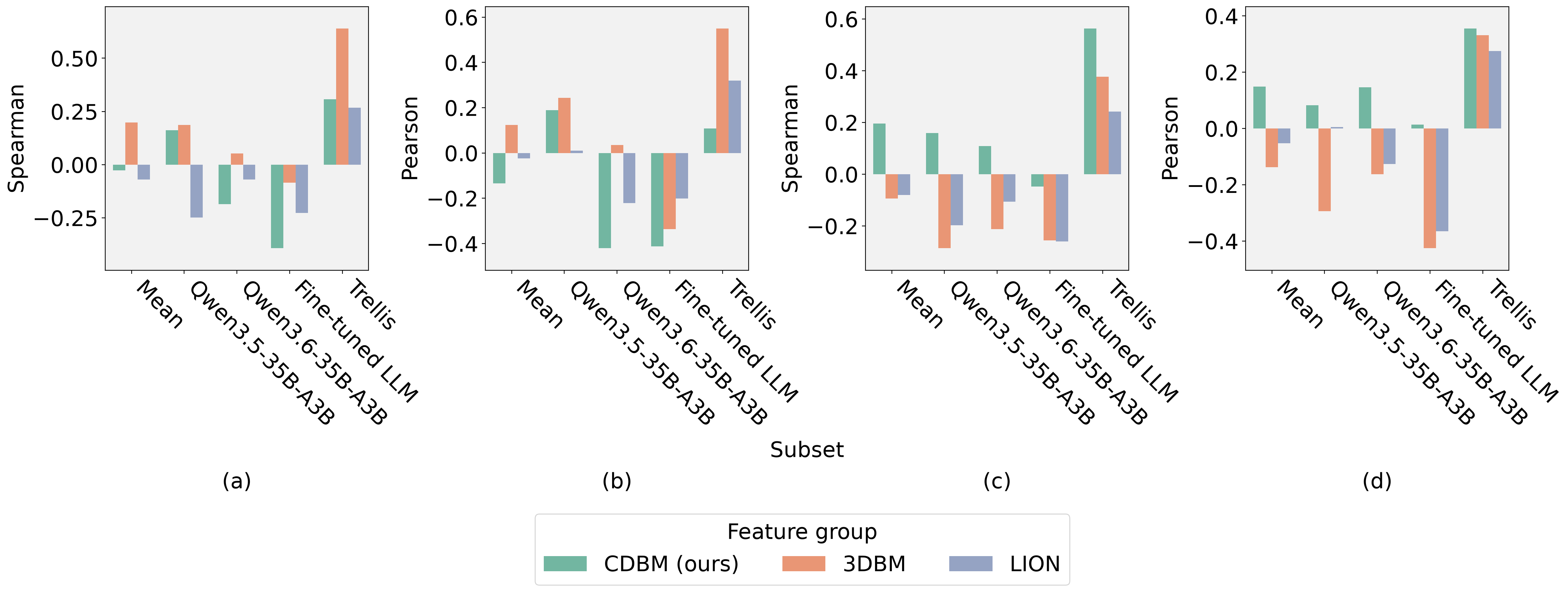}
\caption{Association between classifier realism scores
and human preference scores, evaluated separately
for each generator.
(a)~Spearman and (b)~Pearson correlations for logistic regression;
(c)~Spearman and (d)~Pearson correlations for CatBoost.}
\label{fig:preference}
\end{figure}

\textbf{Main takeaways.} CDBM best separates OSM-derived massings from generated ones with both classifiers. The margin is small for logistic regression (0.862 versus 0.861 for 3DBM) and large for CatBoost (0.972 versus 0.931 for 3DBM and 0.837 for LION). Agreement with human preference depends on the classifier: logistic regression aligns poorly, while CDBM with CatBoost gives positive Spearman correlations for all four generators.

Table~\ref{tab:classifiers_results} compares feature
representations for distinguishing OSM-derived massings
from generated ones.
CDBM achieves the highest score with both classifiers,
although its logistic regression result is nearly
identical to that of 3DBM (0.862 versus 0.861).
The advantage is more pronounced with CatBoost:
CDBM achieves 0.972, compared with 0.931 for 3DBM
and 0.837 for LION.

Figure~\ref{fig:preference} examines whether successful
real--generated discrimination also yields scores
aligned with human visual preferences.
The two left panels show poor alignment for logistic
regression, with 3DBM nevertheless performing best
among the feature representations.
The two right panels show stronger alignment for
CatBoost, particularly when trained on CDBM.
Baseline features produce negative correlations
for most generators, whereas CDBM yields positive
Pearson correlations for all four generator subsets.
Its Spearman correlation is negative for only one subset,
where it is closer to zero than those of the baselines.

\begin{figure}[t]
\centering
\includegraphics[width=0.8\textwidth]{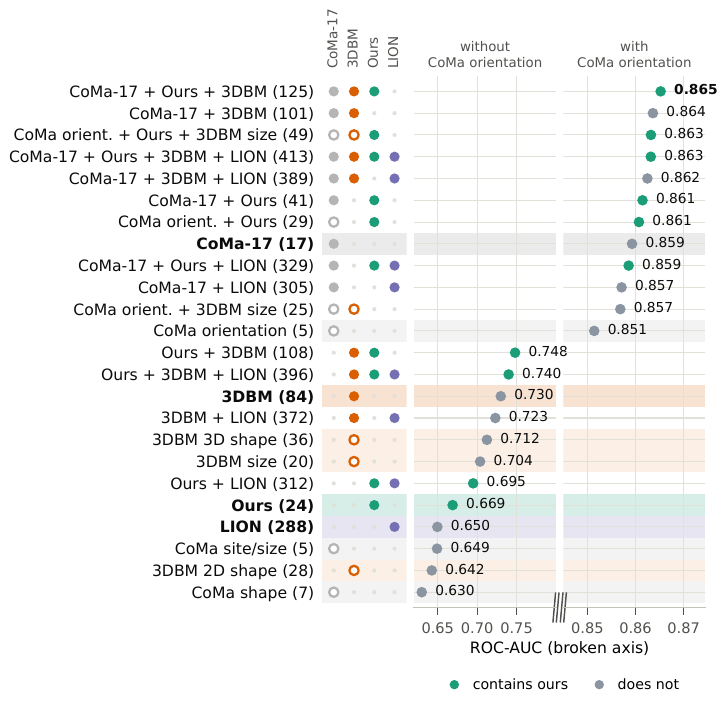}
\caption{ROC-AUC of descriptor sets, averaged over the eight main settings;
feature counts in brackets. Groups are bold and shaded, their families
lighter. Dots on the left show which groups a set contains (filled: the
whole group; ring: part of it). The axis is broken between the sets with and
without CoMa's orientation features. Teal marks sets with our descriptors.}
\label{fig:context_sets}
\end{figure}

These results support CDBM with CatBoost as a proxy
for human visual preference within the evaluated dataset.
The weaker alignment obtained with logistic regression
shows that this relationship depends on how the
descriptor information is combined by the classifier.

\subsection{Contextual Relevance}
\label{sec:context_results}

\textbf{Main takeaways.} Orientation relative to the neighbours dominates contextual relevance: in every setting, each descriptor set that includes it beats each set that does not. Among the four groups, CoMa-17 leads, followed by 3DBM, ours and LION. Our descriptors are complementary: they belong to the best sets and help most where orientation no longer separates true contexts from false ones.

\paragraph{Descriptor sets.}
Orientation relative to the neighbours dominates
Figure~\ref{fig:context_sets} shows ROC-AUC of descriptor sets
for the contextual relevance experiment, averaged over the eight
main settings. Among the four groups, CoMa-17 leads, followed by
3DBM, ours and LION. Orientation relative to the neighbours dominates:
in every setting, each set with CoMa's orientation features beats each
set without them. Ours outperform 3DBM's 2D
indices and CoMa's own shape features, though not 3DBM's 3D-shape and size
blocks. Yet the 19 best of all 255 sets, and the 13 best without
orientation, contain our descriptors, with margins often below 0.001.
The best set, CoMa-17 + Ours + 3DBM, beats CoMa-17 in all 12
settings, and CoMa's five orientation features with ours already exceed
CoMa-17.
Forward selection on training data alone picks our descriptors in all 12
settings and never picks LION.

\begin{figure}[t]
\centering
\includegraphics[width=0.65\textwidth]{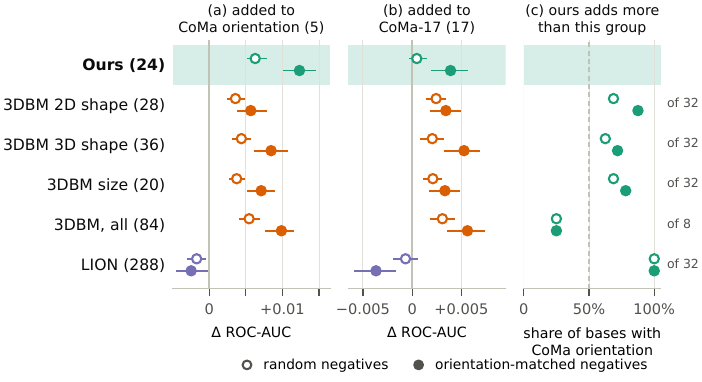}
\caption{What each group adds to sets with CoMa's orientation features, with
random (hollow) and orientation-matched (filled) negatives, each averaged
over its four main settings. (a,~b)~Gain in ROC-AUC on two base sets; bars
are 95\% bootstrap intervals. (c)~Share of the base sets (number on the
right) on which adding our descriptors gains more than adding the group.}
\label{fig:context_additions}
\end{figure}

\paragraph{Where our descriptors help.}
Our descriptors matter most where orientation cannot give the false context
away (Fig.~\ref{fig:context_additions}). With random negatives, a turned
street grid separates most pairs, and ours add almost nothing to CoMa-17,
less than any 3DBM block. With orientation-matched negatives, their gain
roughly doubles on almost every base set: added to CoMa-17, they help as much
as 3DBM's 2D or size block, and added to CoMa's orientation features, more
than any 3DBM block or the full 3DBM. Panel (c) compares the two additions on
every base set with CoMa's orientation features that contains neither of
them: 32 bases for a single 3DBM block or LION, but only 8 for the full 3DBM,
which fills three families. Ours add more than a 3DBM block of similar size
on most of the 32 bases and more than LION on all of them; only the full
3DBM, with 3.5 times as many features, adds more on most of its 8.
Without orientation, almost every addition helps.

\subsection{Geographic Origin and Building Function}
\label{sec:classification_results}

\textbf{Main takeaways.} Adding CDBM to the base features improves test ROC--AUC in all eleven tasks with both classifiers. With the same base, CDBM outperforms LION in all eleven CatBoost tasks and in seven logistic-regression tasks. Across all seven feature sets, 3DBM, with or without the base, is best in 21 of the 22 task--model combinations.

\noindent\textit{Geographic origin.}
The improvement from adding CDBM persists under both
sampling schemes for Global8 and Europe15,
suggesting that its contribution is not restricted
to one sampling configuration.
However, its advantage over LION depends on the
classifier: it holds in all five tasks with CatBoost
and only the two Europe15 tasks with logistic regression.
Base features combined with 3DBM achieve the highest
score in all ten geographic comparisons
(Table~\ref{tab:classification_geography}).

\begin{table}[t]
\caption{Geographic classification:
test macro-averaged one-vs-rest ROC--AUC.
``Only'' denotes the feature representation without
base features; ``+ base'' includes them.
The highest score among all seven feature sets in each
row is bold, and the second-highest is underlined.}
\label{tab:classification_geography}
\centering
\scriptsize
\setlength{\tabcolsep}{2.5pt}
\begin{tabular*}{\linewidth}{
@{\extracolsep{\fill}}llrrrrrrr@{}
}
\toprule
Task & Model & Base
& \multicolumn{2}{c}{CDBM (ours)}
& \multicolumn{2}{c}{LION}
& \multicolumn{2}{c}{3DBM} \\
\cmidrule(lr){4-5}
\cmidrule(lr){6-7}
\cmidrule(lr){8-9}
& &
& Only & + base
& Only & + base
& Only & + base \\
\midrule
\multirow{2}{*}{Global8, location cap}
& CatBoost & 0.7918
& 0.6860 & \underline{0.8365}
& 0.6608 & 0.8069
& 0.8317 & \textbf{0.8739} \\
& LR & 0.7121
& 0.6339 & 0.7323
& 0.6329 & \underline{0.7366}
& 0.7232 & \textbf{0.7750} \\
\midrule
\multirow{2}{*}{Global8, random sampling}
& CatBoost & 0.7669
& 0.6825 & \underline{0.8152}
& 0.6654 & 0.7913
& 0.8087 & \textbf{0.8458} \\
& LR & 0.6911
& 0.6357 & 0.7297
& 0.6374 & \underline{0.7360}
& 0.7168 & \textbf{0.7743} \\
\midrule
\multirow{2}{*}{Urban8}
& CatBoost & 0.9042
& 0.7862 & \underline{0.9456}
& 0.7729 & 0.9231
& 0.9420 & \textbf{0.9594} \\
& LR & 0.7727
& 0.7157 & 0.8205
& 0.7400 & 0.8314
& \underline{0.8367} & \textbf{0.8827} \\
\midrule
\multirow{2}{*}{Europe15, territory balancing}
& CatBoost & 0.7845
& 0.7214 & \underline{0.8521}
& 0.6644 & 0.8024
& 0.8090 & \textbf{0.8566} \\
& LR & 0.7072
& 0.6446 & \underline{0.7479}
& 0.6404 & 0.7387
& 0.7374 & \textbf{0.7963} \\
\midrule
\multirow{2}{*}{Europe15, random sampling}
& CatBoost & 0.7863
& 0.7235 & \underline{0.8529}
& 0.6602 & 0.8050
& 0.8100 & \textbf{0.8567} \\
& LR & 0.7066
& 0.6462 & \underline{0.7474}
& 0.6374 & 0.7364
& 0.7377 & \textbf{0.7961} \\
\bottomrule
\end{tabular*}
\end{table}

\noindent\textit{Building function.}
CDBM improves classification within individual cities
and in both Urban8 samples, where urban-area information
is already available to the classifiers.
The improvement also persists in the smaller samples
of 1{,}800 sites.
Its combination with the base achieves the highest
score among all seven feature sets in the random
Moscow sample with CatBoost; 3DBM leads in the
remaining eleven comparisons
(Table~\ref{tab:classification_function}).

\begin{table}[t]
\caption{Building-function classification:
test macro-averaged one-vs-rest ROC--AUC.
``Only'' denotes the feature representation without
base features; ``+ base'' includes them.
Urban-area information is available to all variants
in the two Urban8 tasks.
The highest score among all seven feature sets in each
row is bold, and the second-highest is underlined.}
\label{tab:classification_function}
\centering
\scriptsize
\setlength{\tabcolsep}{2.5pt}
\begin{tabular*}{\linewidth}{
@{\extracolsep{\fill}}llrrrrrrr@{}
}
\toprule
Task & Model & Base
& \multicolumn{2}{c}{CDBM (ours)}
& \multicolumn{2}{c}{LION}
& \multicolumn{2}{c}{3DBM} \\
\cmidrule(lr){4-5}
\cmidrule(lr){6-7}
\cmidrule(lr){8-9}
& &
& Only & + base
& Only & + base
& Only & + base \\
\midrule
\multirow{2}{*}{Moscow, all eligible}
& CatBoost & 0.8857
& 0.8250 & 0.9357
& 0.8956 & 0.9221
& \textbf{0.9432} & \underline{0.9422} \\
& LR & 0.8546
& 0.7693 & 0.9056
& 0.8840 & 0.9010
& \underline{0.9246} & \textbf{0.9262} \\
\midrule
\multirow{2}{*}{Urban8, all eligible}
& CatBoost & 0.8487
& 0.8087 & 0.8990
& 0.8330 & 0.8660
& \underline{0.9069} & \textbf{0.9085} \\
& LR & 0.7773
& 0.7474 & 0.8239
& 0.8016 & 0.8245
& \underline{0.8585} & \textbf{0.8587} \\
\midrule
\multirow{2}{*}{Urban8, class balancing}
& CatBoost & 0.8546
& 0.8253 & 0.8939
& 0.8435 & 0.8619
& \textbf{0.9020} & \underline{0.9005} \\
& LR & 0.8057
& 0.7993 & 0.8591
& 0.8224 & 0.8347
& \underline{0.8677} & \textbf{0.8685} \\
\midrule
\multirow{2}{*}{Moscow, random sampling}
& CatBoost & 0.8892
& 0.8360 & \textbf{0.9538}
& 0.8543 & 0.8911
& \underline{0.9509} & 0.9498 \\
& LR & 0.8593
& 0.8115 & 0.9141
& 0.8479 & 0.8621
& \textbf{0.9476} & \underline{0.9454} \\
\midrule
\multirow{2}{*}{Calgary, random sampling}
& CatBoost & 0.7958
& 0.7725 & 0.8752
& 0.7943 & 0.8420
& \textbf{0.9031} & \underline{0.8954} \\
& LR & 0.7558
& 0.7326 & 0.8376
& 0.7476 & 0.7687
& \underline{0.8867} & \textbf{0.8963} \\
\midrule
\multirow{2}{*}{Shanghai, random sampling}
& CatBoost & 0.8906
& 0.7458 & 0.9073
& 0.8303 & 0.8891
& \textbf{0.9159} & \underline{0.9132} \\
& LR & 0.7924
& 0.7408 & 0.8466
& 0.8115 & 0.8329
& \underline{0.8733} & \textbf{0.8754} \\
\bottomrule
\end{tabular*}
\end{table}

The representations differ in dimensionality and content:
CDBM contains twelve features, LION 144, and 3DBM 42.
In particular, 3DBM includes additional building-size
measurements that may contribute to classification.
Its predictive advantage therefore cannot be attributed
solely to a more informative description of shape.
CDBM's consistent improvement over the base supports
the value of its compact morphological representation,
while the contribution of feature dimensionality and
additional size information remains unresolved.

\subsection{Geometry-Conditioned Generation Benchmarking}
\label{sec:shape_bench}

\begin{figure}[h]
\includegraphics[width=\textwidth]{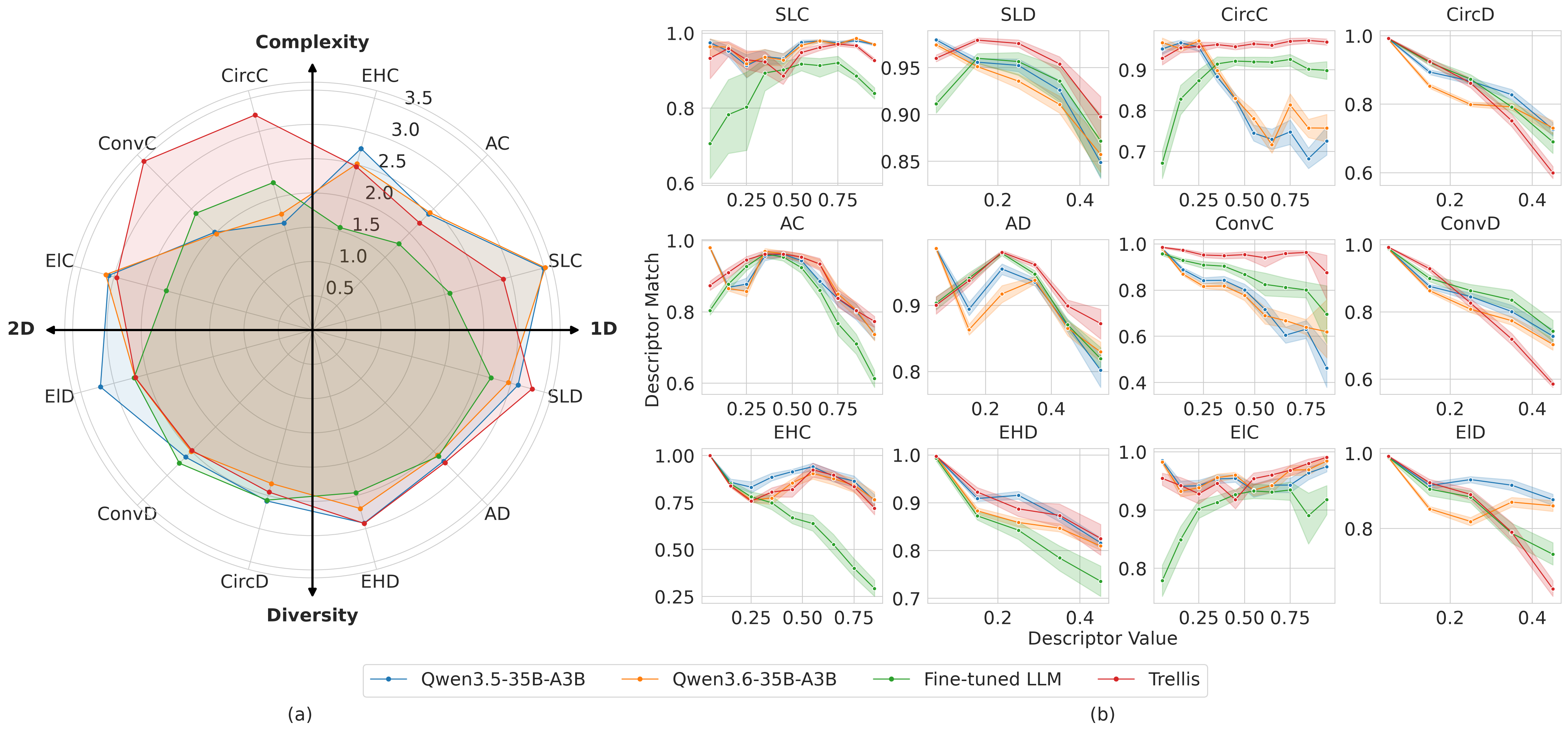}
\caption{Geometry-conditioned generation performance across the twelve CDBM descriptors: a) aggregate target-match scores $s$, displayed as $-\log(1-s)$, b) target-match scores as a function of the requested descriptor value. Colours identify the four evaluated generative models; higher scores indicate closer agreement with the target.}
\label{fig:complexity}
\end{figure}

\textbf{Main takeaways.} No single model controls all twelve descriptors, and scores are lower for complexity than for diversity. In 8 of 12 descriptors, target-match scores fall as the requested value rises, so the models struggle to produce strongly expressed and heterogeneous geometry. Elongation is easier to control than circularity or convexity, particularly for the zero-shot models.

Figure~\ref{fig:complexity}(a) compares the models across
the twelve descriptor-conditioned generation tasks.
Overall, the models obtain lower scores on complexity
tasks than on diversity tasks.
The OSM-trained language and diffusion models achieve
the strongest results on convexity complexity and circularity
complexity, while the zero-shot models lead on the remaining
complexity tasks. For diversity, the diffusion model ranks
first on all three one-dimensional properties: side length,
angles, and extrusion height. No single model consistently
dominates the two-dimensional diversity tasks.
Qwen3.5-35B-A3B outperforms Qwen3.6-35B-A3B on all six
diversity tasks, showing that the newer model does not
necessarily offer better control over these geometric
characteristics.

Figure~\ref{fig:complexity}(b) shows how target-match scores
vary with the requested descriptor value. In 8 of 12
descriptors, scores decline as the target increases,
suggesting that the models struggle to generate strongly
expressed geometric properties and heterogeneous
configurations. This limitation persists despite selecting
the best result from five generation attempts.
However, its extent depends on both the descriptor and
the model.

For circularity diversity and convexity diversity, scores
decrease almost monotonically with increasing target values.
The decline is less pronounced for elongation diversity
in the zero-shot models. On circularity complexity and
convexity complexity, the trained models perform better
at high target values, whereas scores for elongation
complexity are more similar across models.
These observations suggest that elongation may be easier
for the evaluated models to control than circularity
or convexity, particularly in the zero-shot setting.

For side length and angles, the trained language model
exhibits a different pattern: for both complexity and
diversity, its scores peak at intermediate target values
and decrease toward the ends of the range.
This behavior is consistent with a bias toward generating
forms with intermediate descriptor values and limited
responsiveness to more extreme geometric requirements.
It may reflect a preference for common geometric patterns
learned during training, although target-match curves
alone cannot establish memorization of training examples.

\section{Conclusion}
\label{sec:conclusion}

This study introduces CDBM, a set of twelve interpretable descriptors that
measure how strongly the geometric properties of a building massing are
expressed (complexity) and how unevenly they are distributed among its parts
(diversity). Computed for 482 million OpenStreetMap buildings and compared
with 3DBM and LION, the descriptors, used with a non-linear classifier, best
distinguished generated from mapped massings and best aligned with human
preferences. 3DBM remained stronger for building function, region and
contextual relevance, where CDBM nonetheless added complementary
information. Used as generation targets, the descriptors revealed that
current models struggle to produce strongly articulated and heterogeneous
forms.

These results position CDBM as a complement to existing morphological
indicators rather than a replacement. Its value for design lies in
interpretability: instead of a single realism score, it indicates which
property of a generated massing departs from built examples, such as missing
short edges or uniform heights, and allows the same property to be requested
from a generator as a design target. However, its advantages depend on the
classifier and are often small, and the study is limited to single-building
sites at LoD~1.3. Future work will extend the descriptors to multi-building
sites and joint design targets.

\typeout{get arXiv to do 4 passes: Label(s) may have changed. Rerun}
\end{document}